\documentclass[11pt,a4paper]{article}
\usepackage[a-1b]{pdfx}
\usepackage[hyperref]{acl2021}
\usepackage{times}
\usepackage{latexsym}

\usepackage{microtype}

\aclfinalcopy 

\usepackage{todonotes}

\usepackage{enumitem}
\usepackage{booktabs}
\usepackage{multirow}
\usepackage{amsmath}
\usepackage{amsfonts}
\usepackage{amssymb}
\usepackage{fontawesome}
\usepackage[ruled,vlined]{algorithm2e}
\usepackage{arydshln}
\usepackage[procnames]{listings}
\usepackage{multicol}
\usepackage[title]{appendix}
\usepackage{graphicx} 
\usepackage{subcaption}
\usepackage{float}

\title{Modeling Fine-Grained Entity Types with Box Embeddings}

\author{
Yasumasa Onoe$^\spadesuit$,
Michael Boratko$^\diamondsuit$,
Andrew McCallum$^{\diamondsuit \clubsuit}$,
Greg Durrett$^\spadesuit$\\
$^\spadesuit$The University of Texas at Austin \\
$^\diamondsuit$University of Massachusetts Amherst \\
$^\clubsuit$Google Research\\
{\tt\{yasumasa,gdurrett\}@cs.utexas.edu} \\{\tt \{mboratko,mccallum\}@cs.umass.edu} \\{\tt mccallum@google.com}}

\date{}

\begin{document}

\setlength{\abovedisplayskip}{2pt}
\setlength{\belowdisplayskip}{2pt}

\maketitle
\begin{abstract}
Neural entity typing models typically represent fine-grained entity types as vectors in a high-dimensional space, but such spaces are not well-suited to modeling these types' complex interdependencies. We study the ability of \emph{box embeddings}, which embed concepts as $d$-dimensional hyperrectangles, to capture hierarchies of types even when these relationships are not defined explicitly in the ontology. Our model represents both types and entity mentions as boxes. Each mention and its context are fed into a BERT-based model to embed that mention in our box space; essentially, this model leverages typological clues present in the surface text to hypothesize a type representation for the mention. Box containment can then be used to derive both the posterior probability of a mention exhibiting a given type and the conditional probability relations between types themselves. We compare our approach with a vector-based typing model and observe state-of-the-art performance on several entity typing benchmarks. In addition to competitive typing performance, our box-based model shows better performance in prediction consistency (predicting a supertype and a subtype together) and confidence (i.e., calibration), demonstrating that the box-based model captures the latent type hierarchies better than the vector-based model does.\footnote{The code is available at \url{https://github.com/yasumasaonoe/Box4Types}.} 
\end{abstract}

\section{Introduction}\label{sec:intro}
The development of named entity recognition and entity typing has been characterized by a growth in the size and complexity of type sets: from 4 \citep{conll_03} to 17 \citep{Eduard_Hovy_06} to hundreds \cite{bbn, Xiao_Ling_12} or thousands \cite{Eunsol_Choi_18}. These types follow some kind of hierarchical structure \citep{bbn, Xiao_Ling_12, Dan_Gillick_14, Shikhar_Murty_18}, so effective models for these tasks frequently engage with this hierarchy explicitly. Prior systems incorporate this structure via hierarchical losses \cite{Shikhar_Murty_18, Peng_Xu_18, Tongfei_Chen_20} or by embedding types into a high-dimensional Euclidean or hyperbolic space \citep{Dani_Yogatama_15,  Federico_Lopez_20}. However, the former approach requires prior knowledge of the type hierarchy, which is unsuitable for a recent class of large type sets where the hierarchy is not explicit \cite{Eunsol_Choi_18,Yasumasa_Onoe_20}. The latter approaches, while leveraging the inductive bias of hyperbolic space to represent trees, lack a probabilistic interpretation of the embedding and do not naturally capture all of the complex type relationships beyond strict containment.


\begin{figure*}[!t]
    \centering
    \includegraphics[width=1.0\linewidth]{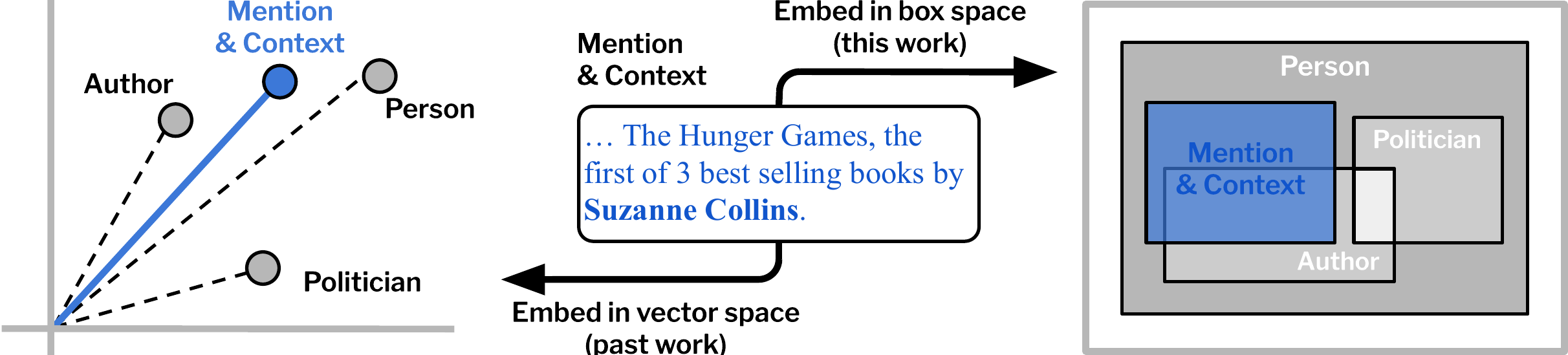}
    \caption{A mention ({\bf Suzanne Collins}) and three entity types are embedded into a vector space (left) and a box space (right). The box space can more richly represent hierarchical interactions between types and uncertainty about the properties of the mention.}
    \label{fig:box_vs_vec}
    \vspace{-15pt}
\end{figure*}



In this paper, we describe an approach that represents entity types with \emph{box embeddings} in a high-dimensional space \cite{Luke_Vilnis_18}. We build an entity typing model that jointly embeds each entity mention and entity types into the same box space to determine the relation between them. Volumes of boxes correspond to probabilities and taking intersections of boxes corresponds to computing joint distributions, which allows us to model mention-type relations (what types does this mention exhibit?) and type-type relations (what is the type hierarchy?). Concretely, we can compute the conditional probability of a type given the entity mention with straightforward volume calculations, allowing us to construct a probabilistic type classification model.

Compared to embedding types as points in Euclidean space \cite{Xiang_Ren_16a}, the box space is expressive and suitable for representing entity types due to its geometric properties. Boxes can nest, overlap, or be completely disjoint to capture subtype, correlation, or disjunction relations, properties which are not explicitly manifested in Euclidean space. The nature of the box computation also allows these complex relations to be represented in a lower-dimensional space than needed by vector-based models.

In our experiments, we focus on comparing our box-based model against a vector-based baseline. We evaluate on four entity typing benchmarks: Ultra-fine Entity Typing \citep{Eunsol_Choi_18}, OntoNotes \citep{Dan_Gillick_14}, BBN \cite{bbn}, and FIGER \citep{Xiao_Ling_12}. To understand the behavior of box embeddings, we further analyze the model outputs in terms of consistency (predicting coherent supertypes and subtypes together), robustness (sensitivity against label noise), and calibration (i.e., model confidence). Lastly, we compare entity representations obtained by the box-based and vector-based models. Our box-based model outperforms the vector-based model on two benchmarks, Ultra-fine Entity Typing and OntoNotes, achieving state-of-the-art-performance. In our other experiments, the box-based model also performs better at predicting supertypes and subtypes consistently and being robust against label noise, indicating that our approach is capable of capturing the latent hierarchical structure in entity types.


\section{Motivation}\label{sec:motivation}

When predicting class labels like entity types that exhibit a hierarchical structure, we naturally want our model's output layer to be sensitive to this structure. Previous work \cite[inter alia]{Xiang_Ren_16a,sonse_shimaoka_17,Eunsol_Choi_18,Yasumasa_Onoe_19} has fundamentally treated types as vectors, as shown in the left half of Figure~\ref{fig:box_vs_vec}. As is standard in multiclass or multi-label classification, the output layer of these models typically involves taking a dot product between a mention embedding and each possible type. A type could be more general and predicted on more examples by having higher norm,\footnote{We do not actually observe this in our vector-based model.} but it is hard for these representations to capture that a coarse type like {\tt Person} will have many mutually orthogonal subtypes.


\begin{figure*}[!t]
    \centering
    \includegraphics[width=1.0\linewidth]{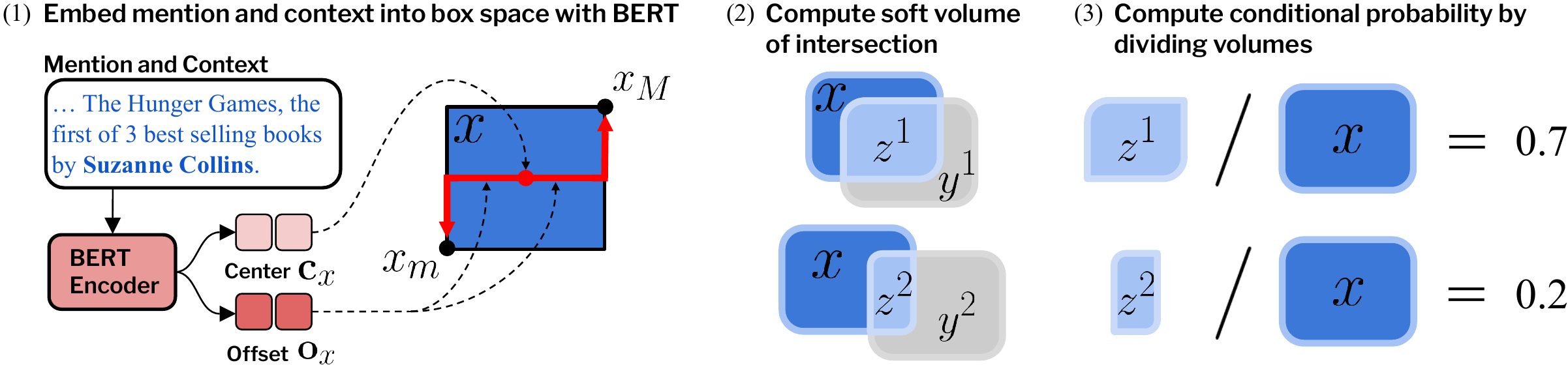}
    \caption{Box-based entity typing model. The mention and context (left) are embedded into the box space and probabilities for each type are computed with a soft volume computation.}
    \label{fig:box_classifier}
    \vspace{-3pt}
\end{figure*}


By contrast, box embeddings naturally represent these kinds of hierarchies as shown in the right half of Figure~\ref{fig:box_vs_vec}. A box that is completely contained in another box is a strict subtype of that box: any entity exhibiting the inner type will exhibit the outer one as well. Overlapping boxes like {\tt Politician} and {\tt Author} represent types that are not related in the type hierarchy but which are not mutually exclusive.
The geometric structure of boxes enables complex interactions with only a moderate number of dimensions \cite{Shib_Sankar_Dasgupta_20}.
\citet{Luke_Vilnis_18}
also define a probability measure over the box space, endowing it with probabilistic semantics. If the boxes are restricted to a unit hypercube, for example, the volumes of type boxes represent priors on types and intersections capture joint probabilities, which can then be used to derive conditional probabilities.

Critically, box embeddings have previously been trained \emph{explicitly} to reproduce a given hierarchy such as WordNet. A central question of this work is whether box embeddings can be extended to model the hierarchies and type relationships that are \textbf{implicit} in entity typing data: we \textbf{do not} assume access to explicit knowledge of a hierarchy during training. While some datasets such as OntoNotes have orderly ontologies, recent work on entity typing has often focused on noisy type sets from crowdworkers \cite{Eunsol_Choi_18} or derived from Wikipedia \cite{Yasumasa_Onoe_20}. We show that box embeddings can learn these structures organically; in fact, they are not restricted to only tree structures, but enable a natural Venn-diagram style of representation for concepts, as with {\tt Politician} and {\tt Author} in Figure~\ref{fig:box_vs_vec}.

\section{Type Modeling with Boxes}
\label{sec:types_with_boxes}

\subsection{Background: Box Embeddings}
\label{sec:background}
Our box embeddings represent entity types as $n$-dimensional hyperrectangles. A box $x$ is characterized by two points $(x_m, x_M)$, where $x_m, x_M \in \mathbb{R}^d$ are the minimum and the maximum corners of the box $x$ and $x_{m, i} \leq x_{M, i}$ for each coordinate $i \in \{1, ..., d\}$. The volume of the box $x$ is computed as $\text{Vol}(x) = \prod_{i} (x_{M, i} - x_{m, i})$. If we normalize the volume of the box space to be $1$, we can interpret the volume of each box as the marginal probability of a mention exhibiting the given entity type. Furthermore, the intersection volume between two boxes, $x$ and $y$, is defined as $\text{Vol}(x \cap y) = \prod_{i} \max\left(\min(x_{M, i}, y_{M, i}) - \max(x_{m, i}, y_{m, i}), 0\right)$ and can be seen as the joint probability of entity types $x$ and $y$. Thus, we can obtain the conditional probability $P (y \mid x) = \frac{\text{Vol}(x \cap y)}{\text{Vol}(x)}$. 

\paragraph{Soft boxes}
Computing conditional probabilities based on hard intersection poses some practical difficulties in the context of machine learning: sparse gradients caused by disjoint or completely contained boxes prevent gradient-based optimization methods from working effectively.
To ensure that gradients always flow for disjoint boxes, \citet{Xiang_Li_19} relax the hard edges of the boxes using Gaussian convolution. We follow the more recent approach of \citet{Shib_Sankar_Dasgupta_20}, who further improve training of box embeddings using max and min Gumbel distributions (i.e., Gumbel boxes) to represent the min and max coordinates of a box.

\subsection{Box-based Multi-label Type Classifier}
\label{sec:box-based-mltc}

Let $s$ denote a sequence of context words and $m$ denote an entity mention span in $s$. Given the input tuple $(m, s)$, the output of the entity typing model is an arbitrary number of predicted types $\{t_0, t_1, ...\}  \in \mathcal{T}$, where $t_k$ is an entity type belonging to a type inventory $\mathcal{T}$. Because we do not assume an explicit type hierarchy, we treat entity typing as a multi-label classification problem, or $|\mathcal{T}|$ independent binary classification problems for each mention. 

Section~\ref{sec:encoder} will describe how to use a BERT-based model to predict a mention and context box\footnote{We could represent mentions as points instead of boxes; however, representing them as boxes enables the size of a mention box to naturally reflect epistemic uncertainty about a mention's types given limited information.} $x$ from $(m, s)$. For now, we assume $x$ is given and we are computing the probability of that mention exhibiting the $k$th entity type, with type box $y^k$. Each type $t^k \in \mathcal{T}$ has a dedicated box $y^k$, which is parameterized by a center vector $\mathbf{c}_y^k \in \mathbb{R}^{d}$ and an offset vector $\mathbf{o}_y^k \in \mathbb{R}^{d}$. The minimum and maximum corners of a box $y^k$ are computed as $y_m^k = \sigma(\mathbf{c}_y^k - \text{softplus}(\mathbf{o}_y^k))$ and $y_M^k = \sigma(\mathbf{c}_y^k + \text{softplus}(\mathbf{o}_y^k))$ respectively, so that parameters $\mathbf{c} \in \mathbb{R}^d$ and $\mathbf{o} \in \mathbb{R}^d$ yield a valid box with nonzero volume.

The conditional probability of the type $t^k$ given the mention and context $(m, s)$ is calculated as
\begin{equation*}
p_\theta(t^k \mid m, s) = \frac{\text{Vol}(z^k)}{\text{Vol}(x)} = \frac{\text{Vol}(x \cap y^k)}{\text{Vol}(x)},
\end{equation*}
where $z^k$ is the intersection between $x$ and $y^k$ ((2) and (3) in Figure~\ref{fig:box_classifier}). Our final type predictions are based on thresholding these probabilities; i.e., predict the type if $p > 0.5$. 

As mentioned in Section~\ref{sec:background}, we use the Gumbel box approach of \citet{Shib_Sankar_Dasgupta_20}, in which the box coordinates are interpreted as the location parameter of a Gumbel max (resp. min) distribution with variance $\beta$. In this approach, the intersection box coordinates become
\begin{equation*} \label{eq:gumbel_min_max}
\begin{aligned}
z_m^k &=  \beta \ln \left(e^{\frac{x_m}{\beta}} + e^{\frac{y^k_m}{\beta}}\right),\\
z_M^k &= - \beta \ln \left(e^{-\frac{x_M}{\beta}} + e^{-\frac{y^k_M}{\beta}}\right).
\end{aligned}
\end{equation*}
Following \citet{Shib_Sankar_Dasgupta_20}, we approximate the expected volume of a Gumbel box using a softplus function:
\begin{equation*}
\text{Vol}(x) \approx \prod_{i} \operatorname{softplus}\left(\frac{x_{M,i} - x_{m,i}}{\beta} - 2 \gamma\right),
\end{equation*}
where $i$ is an index of each coordinate and  $\gamma\approx 0.5772$ is the Euler–Mascheroni constant,\footnote{From \citet{Shib_Sankar_Dasgupta_20}, the Euler-Mascheroni constant appears due to the interpretation of $x_{m,i}, x_{M,i}$ as the \textit{location} parameters of Gumbel distributions.} and $\operatorname{softplus}(x) = \frac{1}{t} \log(1 + \exp(x t))$, with $t$ as an inverse temperature value.

\subsection{Mention and Context Encoder}
\label{sec:encoder}

We format the context words $s$ and the mention span $m$ as $\mathbf{x} =$ {\tt[CLS]} $m$  {\tt[SEP]} $s$ {\tt[SEP]} and chunk into WordPiece tokens \citep{Yonghui_Wu_16}. Using pre-trained BERT\footnote{We use BERT-large uncased (whole word masking) in our experiments.} \citep{Jacob_Devlin_19}, we encode the whole sequence into a single vector by taking the hidden vector at the {\tt [CLS]} token. A highway layer \citep{Rupesh_Kumar_Srivastava_15} projects down the hidden vector $\mathbf{h}^{\texttt{[CLS]}} \in \mathbb{R}^{\ell}$ to the $\mathbb{R}^{2d}$ space, where $\ell$ is the hidden dimension of the encoder (BERT), and $d$ is the dimension of the box space. This highway layer transforms representations in a vector space to the box space without impeding the gradient flow. We further split the hidden vector $\mathbf{\bar h} \in \mathbb{R}^{2d}$ into two vectors: the center point of the box $\mathbf{c}_x \in \mathbb{R}^{d}$ and the offset from the maximum and minimum corners $\mathbf{o}_x \in \mathbb{R}^{d}$. The minimum and maximum corners of the mention and context box are computed as $x_m = \sigma(\mathbf{c}_x - \textsc{Softplus}(\mathbf{o}_x))$ and $x_M = \sigma(\mathbf{c}_x + \textsc{Softplus}(\mathbf{o}_x))$, where $\sigma$ is an element-wise sigmoid function, and $\textsc{Softplus}$ is an element-wise softplus function as defined in Section~\ref{sec:box-based-mltc} ((1) in Figure~\ref{fig:box_classifier}). The output of the softplus is guaranteed to be positive, guaranteeing that the boxes have volume greater than zero.

\subsection{Learning} 
\label{sec:learning}

The goal of training is to find a set of parameters $\theta$ that minimizes the sum of binary cross-entropy losses over all types over all examples in our training dataset $\mathcal{D}$:
\begin{equation*}\label{eq:loss}
\begin{aligned}
\mathcal{L} = - &\sum_{(m, s, \mathbf{t}) \in \mathcal{D}}  \sum_{k} t^k_{\text{gold}} \cdot \log p_\theta(t^k \mid m, s)\\
&+ (1 - t^k_{\text{gold}}) \cdot \log (1 - p_\theta(t^k \mid m, s)),\\
\end{aligned}
\end{equation*}
where $t^k_{\text{gold}} \in \{0, 1\}$ is the gold label for the type $t^k$. We optimize this objective using gradient-based optimization algorithms such as Adam \cite{Kingma_14}.\footnote{With large type sets, most types are highly skewed towards the negative class ($>$99\% negative for many fine-grained types). While past work such as \citet{Eunsol_Choi_18} has used modified training objectives to handle this class imbalance, we did not find any modification to be necessary.}

\section{Experimental Setup}\label{sec:experiments}

Our focus here is to shed light on the difference between type hierarchies learned by the box-based model and the vector-based model. To this end, we first evaluate those two models on standard entity typing datasets. Then, we test models' \emph{consistency}, \emph{robustness}, and \emph{calibration}, and evaluate the predicted types as entity representations on a downstream task (coreference resolution). See Appendix A for hyperparameters.

\subsection{Baseline}

Our chief comparison is between box-based and vector-based modeling of entity types. As our main baseline for all experiments, we use a \textbf{vector-based} version of our entity typing model. We use the same mention and context encoder followed by a highway layer, but this baseline has vector-based type embeddings (i.e., a $| \mathcal{T}|  \times d'$ matrix), and type predictions are given by a dot product between the type embeddings and the mention and context representation followed by element-wise logistic regression. This model is identical to that of \citet{Yasumasa_Onoe_20_Findings} except for the additional highway layer.

\subsection{Evaluation and Datasets}

\paragraph{Entity Typing} We evaluate our approach on the Ultra-Fine Entity Typing (UFET) dataset \citep{Eunsol_Choi_18} with the standard splits (2k for each of train, dev, and test). In addition to the manually annotated training examples, we use the denoised distantly annotated training examples from \citet{Yasumasa_Onoe_19}.\footnote{This consists of 727k training examples derived from the distantly labeled UFET data.} This dataset contains 10,331 entity types, and each type is marked as one of the three classes: \emph{coarse}, \emph{fine}, and \emph{ultra-fine}. Note that this classification \textbf{does not provide explicit hierarchies} in the types, and all classes are treated equally during training.

Additionally, we test our box-based model on three other entity typing benchmarks that have relatively simpler entity type inventories with \textbf{known hierarchies}, namely OntoNotes \citep{Dan_Gillick_14}, BBN \citep{bbn} , and FIGER \citep{Xiao_Ling_12}. See Appendix B for more details on these datasets.

\paragraph{Consistency} A model that captures hierarchical structure should be aware of the relationships between supertypes and subtypes. When a model predicts a subtype, we want it to predict the corresponding supertype together, even when this is not explicitly enforced as a constraint or consistently demonstrated in the data, such as in the UFET dataset. That is, when a model predicts {\tt artist}, {\tt person} should also be predicted. To check this ability, we analyze the model predictions on the UFET dev set. We select 30 subtypes from the UFET type inventory and annotate corresponding supertypes for them in cases where these relationships are clear, based on their cooccurrence in the UFET training set and human intuition. Based on the 30 pairs, we compute accuracy of predicting supertypes and subtypes together. Table~\ref{tab:30types} in Appendix C lists the 30 pairs.  

\paragraph{Robustness} Entity typing datasets with very large ontologies like UFET are noisy; does our box-based model's notion of hierarchy do a better job of handling intrinsic noise in a dataset? To test this in a controlled fashion, we synthetically create noisy labels by randomly dropping the gold labels with probability $\tfrac{1}{3}$.\footnote{If this causes the gold type set to be empty, we retain the original gold type(s); however, this case is rare.} We derive two noisy training sets from the UFET training set: 1) adding noise to the \emph{coarse} types and 2) adding noise to \emph{fine} \& \emph{ultra-fine} types. We train on these noised datasets and evaluate on the standard UFET dev set.

\paragraph{Calibration} \citet{Shrey_Desai_20} study calibration of pre-trained Transformers such as BERT and RoBERTa \citep{Yinhan_Liu_19} on natural language inference, paraphrase detection, and commonsense reasoning. In a similar manner, we investigate if our box-based entity typing model is calibrated: do the probabilities assigned to types by the model match the empirical likelihoods of those types? Since models may naturally have different scales for their logits depending on how long they are trained, we post-hoc calibrate each of our models using temperature scaling \cite{Chuan_Guo_17} and a shift parameter.  We report the total error (e.g., the sum of the errors between the mean confidence and the empirical accuracy) on the UFET dev set and the OntoNotes dev set.

\paragraph{Entity Representations} We are interested in the usefulness of the trained entity typing models in a downstream task. Following \citet{Yasumasa_Onoe_20_Findings}, we evaluate entity representation given by the box-based and vector-based models on the Coreference Arc Prediction (CAP) task \citep{Mingda_Chen_19} derived from PreCo \citep{Hong_Checn_18}. This task is a binary classification problem, requiring to judge if two mention spans (either in one sentence or two sentences) are the same entity or not. As in \citet{Yasumasa_Onoe_20_Findings}, we obtain type predictions (a vector of probabilities associated with types) for each span and use it as an entity representation. The final prediction of coreference for a pair of mentions is given by the cosine similarity between the entity type probability vectors with a threshold $0.5$. The original data split provides 8k examples for each of the training, dev, and test sets. We report accuracy on the CAP test set.

\renewcommand{\arraystretch}{1}
\begin{table}[t]
	\centering
	\small
	\setlength{\tabcolsep}{4pt}
	\begin{tabular}{l  c  c  c }
		\toprule
		\multicolumn{1}{c }{Model}
		 & P & R & F1\\
		\midrule
		Box & 52.8 & \textbf{38.8} & \textbf{44.8}  \\
		Vector & \textbf{53.0} & 36.3 & 43.1  \\
		\midrule
		\citet{Eunsol_Choi_18} & 47.1 & 24.2 & 32.0  \\
		Label GCN \citep{Wenhan_Xiong_19} & 50.3 & 29.2 & 36.9\\
		ELMo \citep{Yasumasa_Onoe_19} & 51.5 & 33.0 & 40.2 \\
		BERT-base \citep{Yasumasa_Onoe_19} & 51.6 & 33.0 & 40.2 \\
		\bottomrule 
	\end{tabular}
	\caption{Macro-averaged P/R/F1 on the test set for the ultra-fine entity typing task of \citet{Eunsol_Choi_18}.}
	\label{tab:ultra-fine-test}
	\vspace{-12pt}
\end{table}

\section{Results and Discussion}\label{sec:results_and_discussion}

\subsection{Entity Typing}
Here we report entity typing performance on Ultra-Fine Entity Typing (UFET), OntoNotes, FIGER, and BBN. For each dataset, we select the best model from 5 runs with different random seeds based on the development performance.

\renewcommand{\arraystretch}{1}
\begin{table*}[t]
	\centering
	\small
	\setlength{\tabcolsep}{4pt}
	\begin{tabular}{l c c c c c c c c c c c c c c c }
		\toprule
		\multicolumn{1}{c}{} & \multicolumn{3}{c}{Total} & \multicolumn{1}{c}{} & \multicolumn{3}{c}{Coarse} & \multicolumn{1}{c}{} & \multicolumn{3}{c}{Fine} & \multicolumn{1}{c}{} & \multicolumn{3}{c}{Ultra-Fine} \\
	    \cmidrule(r){2-4}  \cmidrule(r){6-8} \cmidrule(r){10-12} \cmidrule(r){14-16}
		\multicolumn{1}{c}{Model}
		 & P & R & F1 & & P & R & F1 & & P & R & F1 & & P & R & F1\\
		\midrule
		Box & 52.9 & \textbf{39.1} & \textbf{45.0} & & 71.2 & \textbf{82.5} & \textbf{76.4} & & 50.9 & \textbf{55.2} & \textbf{53.0} & & \textbf{45.4} & \textbf{24.5} & \textbf{31.9} \\
		Vector & \textbf{53.3} & 36.7 & 43.5 & & \textbf{71.7} & 79.9 & 75.6 & & \textbf{51.9} & 48.5 & 50.2 & & 43.7 & 22.7 & 29.8 \\
		\midrule
		\citet{Eunsol_Choi_18} &  48.1 & 23.2 & 31.3 & & 60.3 & 61.6 & 61.0 & & 40.4 & 38.4 & 39.4 & & 42.8 & 8.8 & 14.6\\
		Label GCN \citep{Wenhan_Xiong_19} & 49.3 & 28.1 & 35.8 & & 66.2 & 68.8 & 67.5 & & 43.9 & 40.7 & 42.2 & & 42.4 & 14.2 & 21.3 \\
		ELMo \citep{Yasumasa_Onoe_19} & 50.7 & 33.1 & 40.1 & & 66.9 & 80.7 & 73.2 & & 41.7 & 46.2 & 43.8 & & 45.6 & 17.4 & 25.2\\
		HY XLarge \cite{Federico_Lopez_20} & 43.4 & 34.2 & 38.2 & & 61.4 & 73.9 & 67.1 & & 35.7 & 46.6 & 40.4 & & 36.5 & 19.9 & 25.7 \\
		\bottomrule 
	\end{tabular}
	\caption{Macro-averaged P/R/F1 on the dev set for the entity typing task of \citet{Eunsol_Choi_18} comparing various systems. Our box-based model outperforms models from past work as well as our vector-based baseline.} \label{tab:ultra-fine-dev-breakdown}
	\vspace{-15pt}
\end{table*}

\paragraph{UFET} Table~\ref{tab:ultra-fine-test} shows the macro-precision, recall, and F1 scores on the UFET test set. Our box-based model outperforms the vector-based model and state-of-the-art systems in terms of macro-F1.\footnote{We omit the test number of \citet{Federico_Lopez_20}, since they report results broken down into coarse, fine, and ultra-fine types instead of an aggregated F1 value. However, based on the development results, their approach substantially underperforms the past work of \citet{Yasumasa_Onoe_19} regardless.} Compared to the vector-based model, the box-based model improves primarily in macro-recall compared to macro-precision. \citet{Eunsol_Choi_18} is a LSTM-based model using GloVe \citep{Jeffrey_Pennington_14}. On top of this model, \citet{Wenhan_Xiong_19} add a graph convolution layer to model type dependencies. \citet{Yasumasa_Onoe_19} use ELMo \citep{Matthew_Peters_18} and apply denoising to fix label inconsistency in the distantly annotated data.

Note that past work on this dataset has used BERT-base \cite{Yasumasa_Onoe_19}. Work on other datasets has used ELMo and observed that BERT-based models have surprisingly underperformed \cite{Ying_Lin_19}. Some of the gain from our vector-based model can be attributed to our use of BERT-Large; however, our box model still achieves stronger performance than the corresponding vector-based version which uses the same pre-trained model.

Table~\ref{tab:ultra-fine-dev-breakdown} breaks down the performance into the \emph{coarse}, \emph{fine}, and \emph{ultra-fine} classes. Our box-based model consistently outperforms the vector-based model in macro-recall and F1 across the three classes. The largest gap in macro-recall is in the \emph{fine} class, leading to the largest gap in macro-F1 within the three classes.

We also list the numbers from prior work in Table~\ref{tab:ultra-fine-dev-breakdown}. HY XLarge \citep{Federico_Lopez_20}, a hyperbolic model designed to learn hierarchical structure in entity types, exceeds the performance of the models with similar sizes such as \citet{Eunsol_Choi_18} and \citet{Wenhan_Xiong_19} especially in macro-recall. In the \emph{ultra-fine} class, both our box-based model and HY XLarge achieve higher macro-F1 compared to their vector-based counterparts.

One possible reason for the higher recall of our model is a stronger ability to model dependencies between types. Instead of failing to predict a highly correlated type, the model may be more likely to predict a complete, coherent set of types.

\renewcommand{\arraystretch}{1}
\begin{table*}[t]
	\centering
	\small
	\setlength{\tabcolsep}{4pt}
   \begin{minipage}{1.4\columnwidth}
	\begin{tabular}{l c c c c c c}
		\toprule
		\multicolumn{1}{c}{} & \multicolumn{2}{c}{OntoNotes} & \multicolumn{2}{c}{BBN} & \multicolumn{2}{c}{FIGER} \\
	    \cmidrule(r){2-3}  \cmidrule(r){4-5} \cmidrule(r){6-7}
		\multicolumn{1}{c}{Model}
		 & Ma-F1 & Mi-F1  & Ma-F1 & Mi-F1 & Ma-F1 & Mi-F1 \\
		\midrule
		Box & 77.3 & 70.9  & \:\:78.7\textsuperscript{*} & \:\:78.0\textsuperscript{*}  & 79.4  & 75.0  \\
		Vector & 76.2 & 68.9  & \:\:78.3\textsuperscript{*} & \:\:78.0\textsuperscript{*} & 81.6  & 77.0  \\
		\midrule
		\citet{Sheng_Zhang_18} & 72.1 & 66.5 & 75.7 & 75.1 & 78.7 & 75.5 \\
		\citet{Tongfei_Chen_20} (exclusive) & 72.4 & 67.2 & 63.2 & 61.0 & 82.6 & 80.8 \\
		\citet{Tongfei_Chen_20} (undefined) & 73.0 & 68.1 & 79.7 & 80.5 & 80.5 & 78.1  \\
		\citet{Ying_Lin_19} & \:\:82.9\textsuperscript{\textdagger} & \:\:77.3\textsuperscript{\textdagger} & 79.3 & 78.1 & 83.0 & 79.8  \\
		\bottomrule 
	\end{tabular}
	\caption{Macro-averaged F1 and Micro-averaged F1 on the test set for the entity typing task of OntoNotes, BBN, FIGER. \textdagger: Not directly comparable since large-scale augmented data is used. *: We fix the predictions using simple rules post-hoc.} \label{tab:onto-figer-bbn-test}
	\end{minipage}
	\hspace{4pt}
	\begin{minipage}{.55\columnwidth}
	\begin{tabular}{l c c}
		\toprule
		\multicolumn{1}{c}{} & \multicolumn{1}{c}{BBN} & \multicolumn{1}{c}{FIGER} \\
	    \cmidrule(r){2-2}  \cmidrule(r){3-3} 
		\multicolumn{1}{c}{Model}
		  & Dev Ma-F1 & Dev Ma-F1 \\
		\midrule
		Box & 92.4  & 94.3  \\
		Vector & 92.3  & 94.7  \\
		\bottomrule 
	\end{tabular}
	\caption{Macro-averaged F1 on the dev set of BBN and FIGER. These dev sets are drawn from the same distributions as their training sets.} \label{tab:dev-figer-bbn-test}
	\end{minipage}
	\vspace{-10pt}
\end{table*}

\paragraph{Other datasets} Table~\ref{tab:onto-figer-bbn-test} compares macro-F1 and micro-F1 on the OntoNotes, BBN, and FIGER test sets.\footnote{Note that our hyperparameters are optimized for macro F1 on OntoNotes.} On OntoNotes, our box-based model achieves better performance than the vector-based model. \citet{Sheng_Zhang_18} use document-level information, \citet{Tongfei_Chen_20} apply a hierarchical ranking loss that assumes prior knowledge of type hierarchies, and \citet{Ying_Lin_19} propose an ELMo-based model with an attention layer over mention spans and train their model on the augmented data from \citet{Eunsol_Choi_18}. Among the models trained only on the original OntoNotes training set, the box-based model achieves the highest macro-F1 and micro-F1.

The state-of-the-art system on BBN, the system of \citet{Tongfei_Chen_20} in the ``undefined'' setting, uses explicit knowledge of the type hierarchy. This is particularly relevant on the BBN dataset, where the training data is noisy and features training points with obviously conflicting labels like {\tt person} and {\tt organization}, which appear systematically in the data. To simulate constraints like the ones they use, we use three simple rules to modify our models' prediction: (1) dropping {\tt person} if {\tt organization} exists, (2) dropping {\tt location} if {\tt gpe} exists, and (3) replacing {\tt facility} by {\tt fac}, since both versions of this tag appear in the training set but only {\tt fac} in the dev and test set. Our box-based model and the vector-based model perform similarly and both achieve results comparable with recent systems.

On FIGER, our box-based model shows lower performance compared to the vector-based model, though both are approaching comparable results with state-of-the-art systems. We notice that some of the test examples have inconsistent labels  (e.g., {\tt /organization/sports\_team} is present, but its supertype {\tt /organization} is missing), penalizing models that predict the supertype correctly. In addition, FIGER, like BBN, has systematic shifts between training and test distributions. We hypothesize that our model's hyperparameters (tuned on OntoNotes only) are suboptimal. The high dev performance shown in Table~\ref{tab:dev-figer-bbn-test} implies that our model optimized on held-out training examples may not capture these specific shifts as well as other models whose inductive biases are better suited to this unusually mislabeled data.

\subsection{Consistency}

One factor we can investigate is whether our model is able to predict type relations in a sensible, consistent fashion \emph{independent of the ground truth for a particular example}. For this evaluation, we investigate our model's predictions on the UFET dev set. We count the number of occurrences for each subtype in 30 supertype/subtype pairs (see Table~\ref{tab:30types} in Appendix C). Then, for each subtype, we count how many times its corresponding supertype is also predicted. Although these supertype-subtype relations are not strictly defined in the training data, we believe they should nevertheless be exhibited by models' predictions. Accuracy is given by the ratio between those counts, indicating how often the supertype was correctly picked up.

Table~\ref{tab:consistency} lists the total and per-supertype accuracy on the supertype/subtype pairs. We report the number of subtypes grouped by their supertypes to show their frequency (the ``Count'' column in Table~\ref{tab:consistency}). Our box-based model achieves better accuracy compared to the vector-based model on all supertypes. The gaps are particularly large on {\tt place} and {\tt organization}. Note that some of the UFET training examples have inconsistent labels (e.g., a subtype {\tt team} can be a supertype {\tt organization} or {\tt group}), and this ambiguity potentially confuses a model during training. Even in those tricky cases, the box-based model shows reasonable performance. The geometry of the box space itself gives some evidence as to why this consistency would arise (see Section~\ref{sec:box_overlaps} for visualization of box edges).

\subsection{Robustness}
Table~\ref{tab:noised_ufet} analyzes models' sensitivity to the label noise. We list the UFET dev performance by models trained on the noised UFET training set. When the \emph{coarse} types are noised (i.e., omitting some supertypes), the vector-based model loses $4.8$ points of macro-F1 while our box-based model only loses $1.5$ points. A similar trend can be seen when the \emph{fine} and \emph{ultra-fine} types are noised (i.e., omitting some subtypes). In both cases, the vector-based model shows lower recall compared to the same model trained on the clean data, while our box-based model is more robust. We also note that the vector-based model tends to overfit to the training data quickly. We hypothesize that the use of boxes works as a form of regularization, since moving boxes may be harder than moving points in a space, thus being less impacted by noisy labels.

\renewcommand{\arraystretch}{1}
\begin{table}[t]
	\centering
	\small
	\setlength{\tabcolsep}{4pt}
	\begin{tabular}{l c c c c}
		\toprule
		\multicolumn{1}{c}{} & \multicolumn{2}{c}{Box} & \multicolumn{2}{c}{Vector}\\
		\cmidrule(r){2-3}  \cmidrule(r){4-5}
		\multicolumn{1}{c}{Supertype} & \multicolumn{1}{c}{Count} & \multicolumn{1}{c}{Acc.} & \multicolumn{1}{c}{Count} & \multicolumn{1}{c}{Acc.}\\
		\midrule
		 {\tt person} & 982 & \textbf{99.7} & 745 & 98.6\\
		 {\tt location} & 470 & \textbf{86.1} & 450 & 84.4\\
		 {\tt place} & 49 & \textbf{95.9} & 29 & 68.9 \\
		 {\tt organization} & 496 & \textbf{84.6} & 407 & 77.8 \\
		 \midrule
		 Total & 1,997 & \textbf{92.7} & 1,631 & 89.0\\
		\bottomrule 
	\end{tabular}
	\caption{Consistency: accuracy evaluated on the 30 supertype \& subtypes pairs. The ``Count'' column shows the number of subtypes found in the predictions. The accuracy is the frequency of predicting the corresponding supertype when the subtype is exhibited.}
	\label{tab:consistency}
	\vspace{-0pt}
\end{table}

\renewcommand{\arraystretch}{1}
\begin{table}[t]
	\centering
	\small
	\setlength{\tabcolsep}{4pt}
	\begin{tabular}{l l c  c  c  c}
		\toprule
		\multicolumn{1}{c}{Training Data} & \multicolumn{1}{c}{Model}& \multicolumn{1}{c}{P} & \multicolumn{1}{c}{R} & \multicolumn{1}{c}{F1}  & \multicolumn{1}{c}{$\Delta$ in F1} \\
		\midrule
		 \multirow{2}{*}{Noised Coarse} & Box & 51.0 & \textbf{37.9} & \textbf{43.5} & -1.5 \\
		 & Vector & \textbf{51.5} & 31.0 & 38.7 & -4.8\\
		 \midrule
		 Noised Fine & Box & 53.0 & \textbf{37.2} & \textbf{43.7} & -1.3 \\
		 \:\:\:\:\& Ultra-fine& Vector & \textbf{58.6} & 30.6 & 40.2 & -3.3\\
		\bottomrule 
	\end{tabular}
	\caption{Entity typing results of the UFET dev set. Models are trained on the noised UFET training set. The ``$\Delta$ in F1'' column shows the performance drop from the model trained on the original UFET training set (not noised).}
	\label{tab:noised_ufet}
	\vspace{-4pt}
\end{table}

\renewcommand{\arraystretch}{1}
\begin{table}[t]
	\small
	\setlength{\tabcolsep}{4pt}
	\begin{minipage}{.6\columnwidth}
	 \centering
	\begin{tabular}{l  c  c}
		\toprule
		\multicolumn{1}{l}{Model} & \multicolumn{1}{l}{Scale / Shift} & \multicolumn{1}{l}{Total Error} \\
		 \midrule
 		 \multicolumn{3}{c}{UFET} \\
		 \midrule
		 Box & 0.5 / -1.1 & \textbf{0.1119}\\
         Vector & 0.2 / -1.1 &  0.3279 \\
		 \midrule
 		 \multicolumn{3}{c}{OntoNotes}\\
		 \midrule
		 Box & 0.9 / -0.3 & \textbf{0.1358}\\
         Vector & 0.7 / -0.4 & 0.1568\\
		\bottomrule 
	\end{tabular}
	\caption{Total calibration error on UFET and OntroNotes. We scale and shift logits post-hoc.}
	\label{tab:calibration}
	\end{minipage}
	\hspace{3pt}
	\begin{minipage}{.34\columnwidth}
	\vspace{-23pt}
	 \centering
		\begin{tabular}{l c}
		\toprule
		\multicolumn{1}{l}{Model} & {Test Acc.}\\
		\midrule
        Box & \textbf{78.1} \\
        Vector &  77.3 \\
        Random & 50.0 \\
		\bottomrule
	\end{tabular}
	\caption{Accuracy on the CAP test set \cite{Mingda_Chen_19}. This is a binary classification task.}
	\label{tab:cap}
	\end{minipage}
	\vspace{-6pt}
\end{table}

\subsection{Calibration}

Following \citet{Khanh_Nguyen_15}, we split model confidence (output probability) for each typing decision of each example into 10 bins (e.g., 0-0.1, 0.1-0.2 etc.). For each bin, we compute mean confidence and empirical accuracy. We show the total calibration error (lower is better) as well as the scaling and shifting constants in Table~\ref{tab:calibration}. As the results on UFET and OntoNotes show, both box-based and vector-based entity typing models can be reasonably well calibrated after applying temperature scaling and shifting. However, the box-based model achieves slightly lower total error. 

\subsection{Entity Representation for Coreference}
This experiment evaluates if model outputs are immediately useful in a downstream task. For this task, we use the box-based and vector-based entity typing models trained on the UFET training set (i.e., we do not train models on the CAP training set). Table~\ref{tab:cap} shows the test accuracy on the CAP data. Our box-based model achieves slightly higher accuracy than the vector-based model, indicating that ``out-of-the-box'' entity representations obtained by the box-based model contains more useful features for the CAP task.\footnote{Our results are not directly comparable to those of \citet{Yasumasa_Onoe_20_Findings}; we train on the training set of UFET dataset, and they train on examples from the train, dev, and test sets.}

\begin{figure*}[t!]
\centering
\begin{subfigure}[H]{\linewidth}
\centering
  \includegraphics[width=\linewidth]{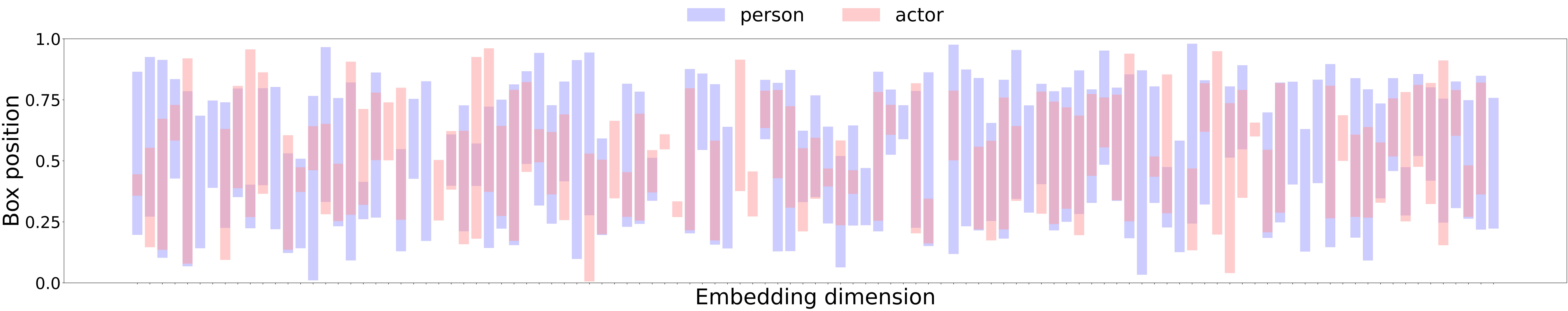}
  \vspace{-16pt}
  \caption{}
\end{subfigure}\\
\vspace{6pt}
\begin{subfigure}[H]{\linewidth}
\centering
   \includegraphics[width=\linewidth]{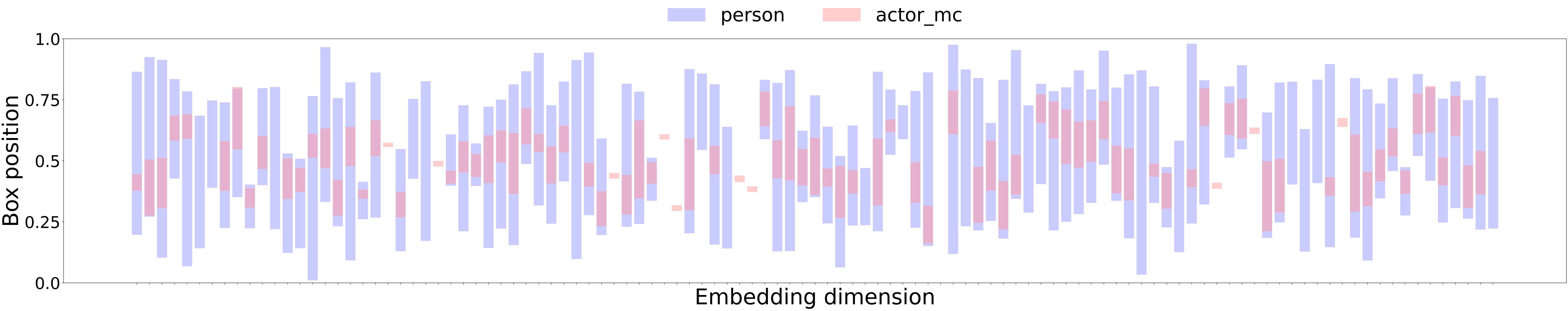}
   \vspace{-16pt}
   \caption{}
\end{subfigure}\\
\vspace{-6pt}
\caption{Edges of (a) the {\tt person} box vs the {\tt actor} box and (b) the {\tt person} box vs the minimum bounding box of the intersections between mention \& context boxes and the {\tt actor} box.}
\label{fig:edges}
\vspace{-8pt}
\end{figure*}

\subsection{Box Edges}\label{sec:box_overlaps}
To analyze how semantically related type boxes are located relative to one another in the box space, we plot the edges of the {\tt person} and {\tt actor} boxes along the 109 dimensions one by one. Figure~\ref{fig:edges} shows how those two boxes overlap each other in the high-dimensional box space. The upper plot in Figure~\ref{fig:edges} compares the {\tt person} box and the {\tt actor} box learned on the UFET data. We can see that the edges of {\tt person} contain the edges of {\tt actor} in many dimensions but not all, meaning that the {\tt person} box overlaps with the {\tt actor} box but doesn't contain it perfectly as we might expect.

However, we can additionally investigate whether the  {\tt actor} box is \emph{effectively} contained in the  {\tt person} for parts of the space actually used by the mention boxes. The lower plot in Figure~\ref{fig:edges} compares the {\tt person} box and the minimum bounding box of the intersections between the {\tt actor} and the mention and context boxes obtained using the UFET dev examples where the {\tt actor} type is predicted. This minimum bounding box approximates the effective region within the {\tt actor} box. Now the edges of {\tt actor} are contained in the edges of {\tt person} in the most of dimensions, indicating that the {\tt person} box almost contains this ``effective'' {\tt actor} box.

\section{Related Work}

\paragraph{Embeddings} Embedding concepts/words into a high-dimensional vector space \citep{Hinton_86} has a long history and has been an essential part of neural networks for language \citep{Yoshua_Bengio_03, Ronan_Collobert_11}.   
There is similarly a long history of rethinking the semantics of these embedding spaces, such as treating words as regions using sparse count-based vectors \citep{Katrin_Erk_09_b,Katrin_Erk_09_a} or dense distributed vectors \citep{Luke_Vilnis_15}. Order embeddings \citep{Ivan_Vendrov_16} or their probabilistic version (POE) \citep{Alice_Lai_17} are one technique suited for hierarchical modeling. However, OE can only handle binary entailment decisions, and POE cannot model negative correlations between types, a critical limitation in its use as a probabilistic model; these shortcomings directly led to the development of box embeddings. Hyperbolic embeddings \cite{NickelKiela2017,Federico_Lopez_20} can also model hierarchical relationships as can hyperbolic entailment cones \cite{Octavian_Eugen_Ganea_18}; however, these approaches lack a probabilistic interpretation.

Recent work on knowledge base completion \citep{Ralph_Abboud_20} and reasoning over knowledge graphs \citep{Hongyu_Ren_20} embeds relations or queries using box embeddings, but entities are still represented as vectors. In contrast, our model embed both entity mentions and types as boxes.

\paragraph{Entity typing} Entity typing and named entity recognition \cite{conll_03} are old problems in NLP. Recent work has focused chiefly on predicted fine-grained entity types \cite{Xiao_Ling_12,Dan_Gillick_14,Eunsol_Choi_18}, as these convey significantly more information for downstream tasks. As a result, there is a challenge of scaling to large type inventories, which has inspired work on type embeddings \cite{Xiang_Ren_16a, Xiang_Ren_16b}.

Entity typing information has been used across a range of NLP tasks, including models for entity linking and coreference \citep{Durrett_Klein_14}. Typing has been shown to be useful for cross-domain entity linking specifically \citep{Nitish_Gupta_17,Yasumasa_Onoe_20}. It has also recently been applied to coreference resolution \cite{Yasumasa_Onoe_20_Findings,Khosla_Rose_2020_Type} and text generation \cite{Dong_2020_Injecting}, suggesting that it can be a useful intermediate layer even in pre-trained neural models.

\section{Conclusion}\label{conclusion}

In this paper, we investigated a box-based model for fine-grained entity typing. By representing entity types in a box embedding space and projecting entity mentions into the same space, we can naturally capture the hierarchy of and correlations between entity types. Our experiments showed several benefits of box embeddings over the equivalent vector-based model, including typing performance, calibration, and robustness to noise.

\section*{Acknowledgments}

Thanks to the members of the UT TAUR lab, Pengxiang Cheng, and Eunsol Choi for helpful discussion; Tongfei Chen and Ying Lin for providing the details of experiments.
This work was also partially supported by NSF Grant IIS-1814522, NSF Grant SHF-1762299, and based on research in part supported by the Air Force Research Laboratory (AFRL), DARPA, for the KAIROS program under agreement number FA8750-19-2-1003, as well as
University of Southern California subcontract no.~123875727 under Office of Naval Research prime contract no.~N660011924032.
The U.S. Government is authorized to reproduce and distribute reprints for Governmental purposes notwithstanding any copyright notation thereon. The views and conclusions contained herein are those of the authors and should not be interpreted as necessarily representing the official policies or endorsements, either expressed or implied, of AFRL, DARPA, or the U.S. Government.

\bibliography{acl2021}
\bibliographystyle{acl_natbib}

\newpage
\appendix

\section*{Appendix A: Hyperparameter Search}\label{app:hyperparams}

We use Bayesian hyperparameter tuning and the Hyperband stopping criteria \citep{Lisha_Li_17} implemented in the Weights \& Biases software \citep{wandb}. We use Adam \citep{Kingma_14} for all experiments.  We perform hyperparameter search on OntoNotes due to its fast convergence. This finds a lower dimension for the box-based model compared to the vector-based model (109-$d$ $\times$ 2 vs 307-$d$), resulting fewer parameters in the box-based model. When we train the box-based model on the UFET dataset, we sample 1,000 negatives (i.e., wrong types) to speed up convergence; this is not effective in the vector-based model, so we do not do this there.

We use the same hyperparameters for the other three datasets. We train all models using NVIDIA V100 GPU with batch size $128$. We implement our models using HuggingFace's Transformers library \citep{Thomas_Wolf_20}.

Table~\ref{tab:hyperparams} shows hyperparameters of the box-based and vector-based models as well as their ranges to search. For Adam, we use $\beta_1 = 0.9$ and $\beta_2 = 0.999$ for training. 

\renewcommand{\arraystretch}{1}
\begin{table}[H]
	\centering
	\small
	\setlength{\tabcolsep}{4pt}
	\begin{tabular}{l l c c}
		\toprule
		\multicolumn{1}{c}{Model} & \multicolumn{1}{c}{Hyperparameter} & \multicolumn{1}{c}{Range} & \multicolumn{1}{c}{Selected}\\
		\midrule
		\multirow{6}{*}{Box} & Batch Size & \{16, 32, 64, 128\} & 128 \\
		& lr (BERT) & - & 2e-5 \\
		& lr (Other) & [0.0001, 0.01] &  0.00372 \\
		& Box Dimension & [50, 250] & 109 \\		
		& Gumbel Temp. & [0.0001, 0.01]\textsuperscript{*} & 0.00036 \\
		& Softplus Temp.\textsuperscript{\textdagger} & [0.1, 10]\textsuperscript{*} & 1.2471 \\
		\midrule
		\multirow{4}{*}{Vector} & Batch size & \{16, 32, 64, 128\} & 128 \\
		& lr (BERT) & - & 2e-5 \\
		& lr (Other) & [0.0001, 0.01] & 0.00539  \\
		& Vector Dimension & [100, 500] & 307 \\
		\bottomrule
	\end{tabular}

	\caption{Hyperparameters and their ranges. *: we use a log uniform distribution. \textdagger: Pytorch implementation of a softplus function takes inverse $\beta$. }
	\label{tab:hyperparams}
\end{table}

\section*{Appendix B: Entity Typing Benchmarks}\label{app:typing_data}
OntoNotes \citep{Dan_Gillick_14} has 89 types with a 3-level hierarchy (e.g., {\tt /location/geography/mountain}). We use the same splits (250k train / 2k dev / 9k test) provided by \citep{sonse_shimaoka_17}. FIGER \citep{Xiao_Ling_12}, derived from Wikipedia, uses 113 types with a 2-level hierarchy (e.g., {\tt /person/musician}). We use the same splits (2M train / 10k dev / 563 test) as \citep{sonse_shimaoka_17}. BBN \citep{bbn} is based on the one million word Penn Treebank corpus from Wall Street Journal articles. We use the same splits (84k train / 2k dev / 14k test) as \citet{Xiang_Ren_16b, Tongfei_Chen_20}.

\section*{Appendix C: Supertype/subtype pairs}\label{app:30types}

Table~\ref{tab:30types} shows the supertype/subtype pairs we manually annotated for our consistency test.

\renewcommand{\arraystretch}{1}
\begin{table}[H]
	\centering
	\small
	\setlength{\tabcolsep}{4pt}
	\begin{tabular}{c c}
		\toprule
		\multicolumn{1}{c}{Supertype} & \multicolumn{1}{c}{Subtype}\\
		\midrule
        {\tt person} & {\tt politician} \\
        {\tt person} & {\tt athlete} \\
        {\tt person} & {\tt leader} \\
        {\tt person} & {\tt official} \\
        {\tt person} & {\tt spokesperson} \\
        {\tt person} & {\tt musician} \\
        {\tt person} & {\tt actor} \\
        {\tt person} & {\tt professional} \\
        {\tt person} & {\tt male} \\
        {\tt person} & {\tt female} \\
        {\tt location} & {\tt country} \\
        {\tt location} & {\tt city} \\
        {\tt location} & {\tt area} \\
        {\tt location} & {\tt region} \\
        {\tt location} & {\tt position} \\
        {\tt location} & {\tt space} \\
        {\tt location} & {\tt district} \\
        {\tt location} & {\tt territory} \\
        {\tt place} & {\tt structure} \\
        {\tt place} & {\tt building} \\
        {\tt organization} & {\tt company} \\
        {\tt organization} & {\tt institution} \\
        {\tt organization} & {\tt government} \\
        {\tt organization} & {\tt agency} \\
        {\tt organization} & {\tt team} \\
        {\tt organization} & {\tt administration} \\
        {\tt organization} & {\tt military} \\
        {\tt organization} & {\tt association} \\
        {\tt organization} & {\tt social\_group} \\
        {\tt organization} & {\tt committee} \\
		\bottomrule
	\end{tabular}
	\caption{30 supertype and subtype pairs used for the consistency test.}
	\label{tab:30types}
\end{table}

\begin{figure*}[t!]
\centering
\begin{subfigure}[H]{\linewidth}
  \includegraphics[width=\linewidth]{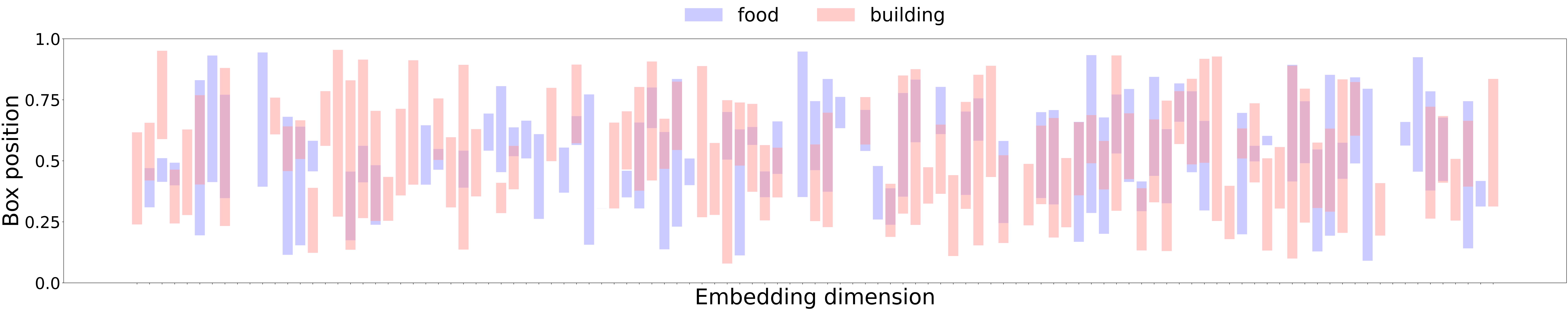}
  \vspace{-16pt}
  \caption{}
\end{subfigure}\\
\vspace{6pt}
\begin{subfigure}[H]{\linewidth}
  \includegraphics[width=\linewidth]{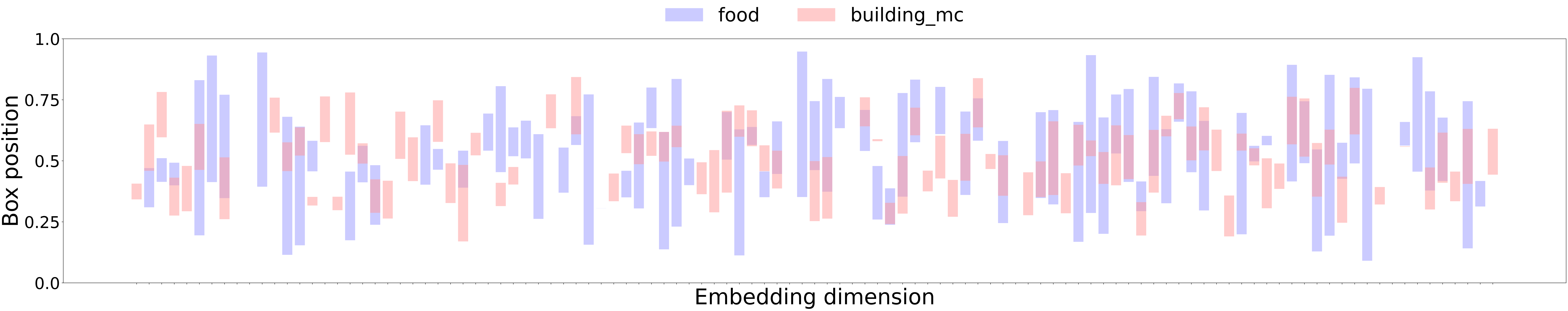}
  \vspace{-16pt}
  \caption{}
\end{subfigure}
\vspace{-6pt}
\caption{Edges of (a) the {\tt food} box vs the {\tt building} box and (b) the {\tt food} box vs the minimum bounding box of the intersections between mention \& context boxes and the {\tt building} box.}
\label{fig:edges2}
\end{figure*}

\begin{figure*}[t]
\centering
\begin{subfigure}[H]{\columnwidth}
\centering
   \includegraphics[width=0.8\columnwidth]{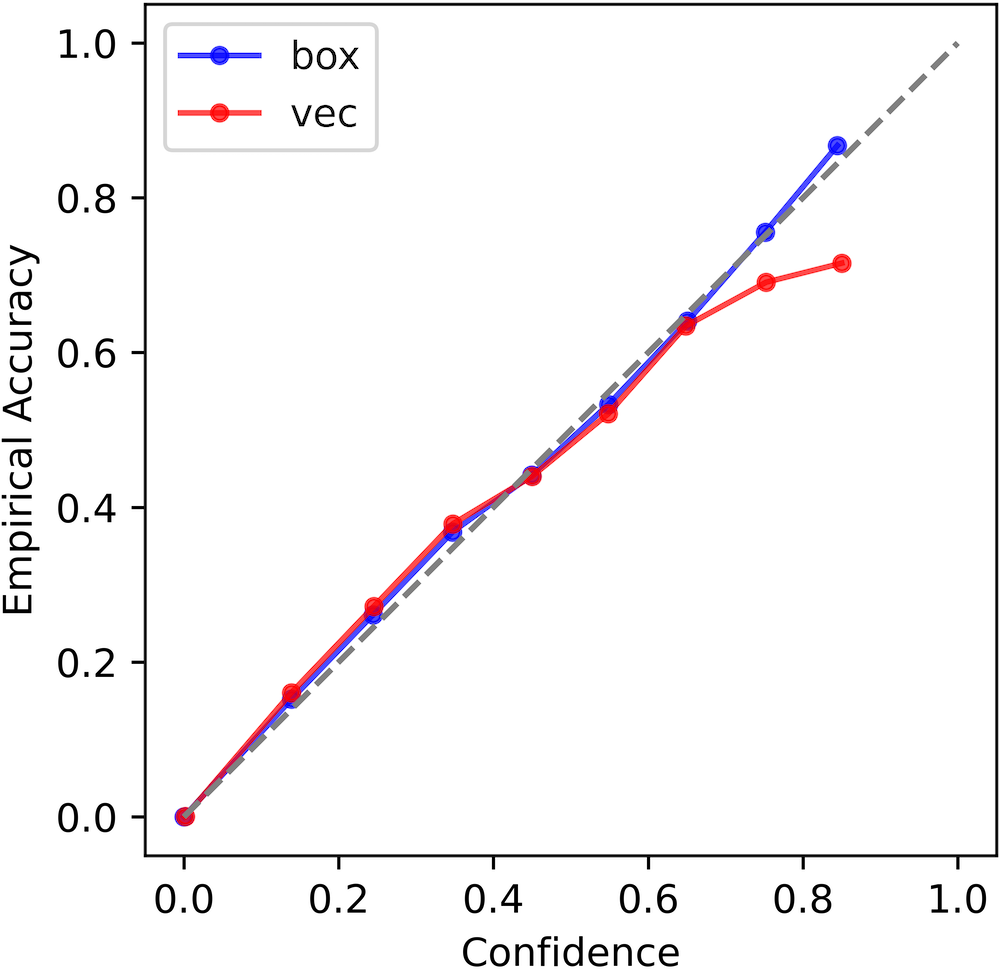}
   \caption{UFET}
\end{subfigure}
\begin{subfigure}[H]{\columnwidth}
\centering
   \includegraphics[width=0.8\columnwidth]{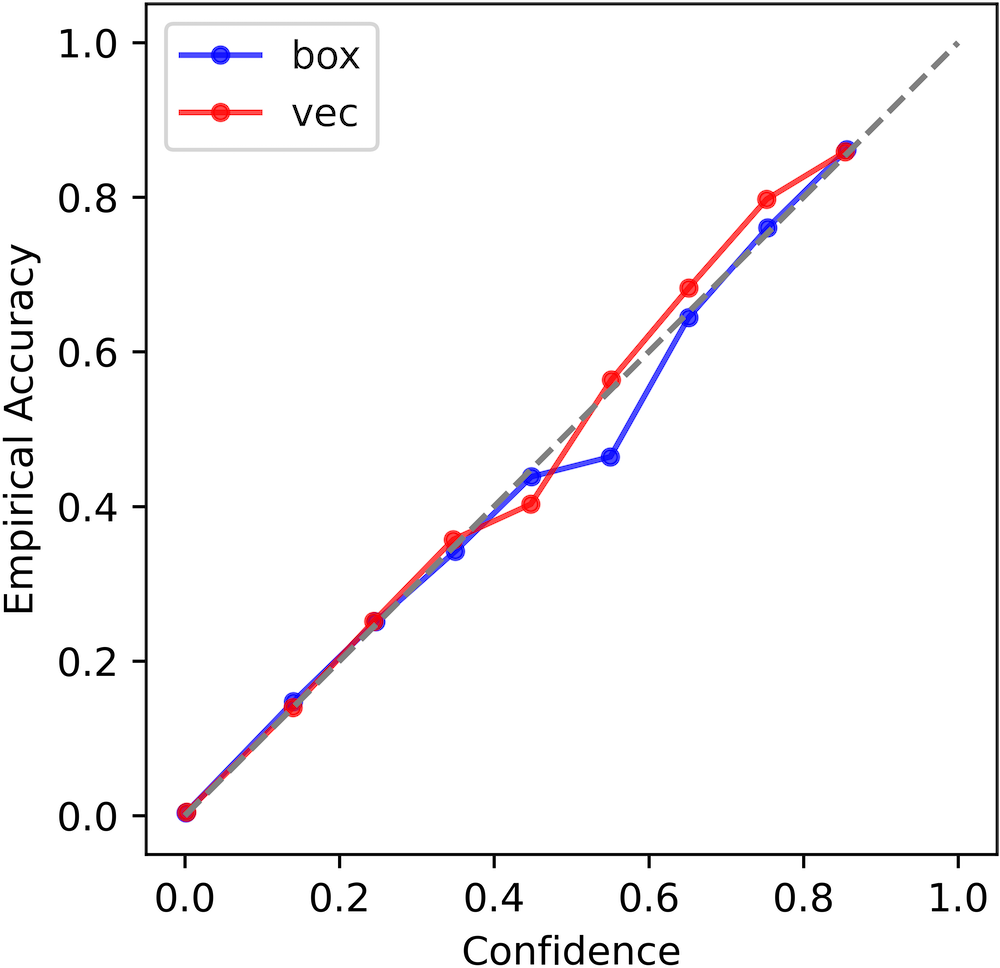}
   \caption{OntoNotes}
\end{subfigure}
\caption{Reliability Plots on (a) UFET and (b) OntoNotes.}
\label{fig:calibration}
\end{figure*}

\section*{Appendix D: Box Edges}\label{app:box_overlaps}

Similar to Figure~\ref{fig:edges}, we plot the semantically unrelated type boxes {\tt food} and {\tt building} in Figure~\ref{fig:edges2}. These boxes are largely misaligned as expected, and the minimum bounding box of the intersections between the {\tt building} and the mention and context boxes is also off from the {\tt food} box.

\section*{Appendix E: Reliability Plot}\label{app:calibration}
Figure~\ref{fig:calibration} visualizes the alignment between confidence and empirical accuracy on the UFET and OntoNotes dev sets.

\end{document}